\documentclass{article}

 \usepackage[preprint]{neurips_2026}

\usepackage[utf8]{inputenc} 
\usepackage[T1]{fontenc}    
\usepackage{hyperref}       
\usepackage{url}            
\usepackage{booktabs}       
\usepackage{amsfonts}       
\usepackage{nicefrac}       
\usepackage{microtype}      
\usepackage[table,xcdraw,dvipsnames]{xcolor}  
\usepackage{graphicx}
\usepackage{xspace}
\usepackage{wrapfig}
\usepackage{multirow}
\usepackage{amsmath}
\usepackage{amssymb}
\usepackage{enumitem}
\usepackage{algorithm}
\usepackage{algorithmic}
\usepackage{bm}
\usepackage[most]{tcolorbox}

\newcommand{\gainvalue}[3]{\hspace{#3mm}#1\scalebox{0.7}{\textcolor[HTML]{a6192e}{{+#2}}}}
\newcommand{\nonevalue}[3]{\hspace{#3mm}#1\scalebox{0.7}{\textcolor{gray}{{+#2}}}}

\newcommand{\github}{\raisebox{-1.7pt}{\includegraphics[height=1.05em]{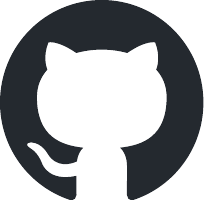}}\xspace}
\title{Not only where, But when: Temporal Scheduling for RLVR}

\author{
Jinghao Zhang\,$^{1,2,4}$ \quad 
Ruilin Li\,$^{2,3,4}$ \quad 
\textbf{Feng Zhao}\,$^{1,\dagger}$ \quad 
\textbf{Jiaqi Wang}\,$^{2,4,\dagger}$ \\
$^1$University of Science and Technology of China \quad 
$^2$Shanghai Innovation Institute \\
$^3$Wuhan University \quad 
$^4$JD.com \\
\vspace{0.3em}
\github \href{https://github.com/Jinghaoleven/RLVR-Schedule}{\texttt{https://github.com/Jinghaoleven/RLVR-Schedule}}
}

\begin{document}

\maketitle

\begin{abstract}
Reinforcement learning with verifiable rewards (RLVR) has become a core technique for post-training of Large Language Models (LLMs). 
While policy optimization is driven by all sampled tokens under a globally broadcast scalar reward, the heterogeneous policy behaviors exhibited along trajectories are largely overlooked without differentiation.
Existing works address this by credit allocation, including token-level advantage reweighting, and selective token optimization, however, the allocation criterion are principally stagnant throughout training, limiting resilient policy evolution.
In this work, we argue that \textit{when} learning signals are scheduled can be as important as \textit{where} they are allocated across tokens, and introduce the temporal dimension that scheduling the credit allocation criteria over the course of RLVR optimization.
We find that prioritizing targeted tokens emphasized with specific policy behaviors, and gradually attenuating toward general optimization leads to more stable and efficient learning dynamics.
Furthermore, we show that simple trajectory percentiles provide a natural perspective for distinguishing policy behaviors, and works effectively with temporal scheduling.
Our analysis reveals that standard optimization substantially sacrifices policy entropy when simultaneously accommodating heterogeneous behaviors, whereas temporal scheduling yields healthier policy evolution dynamics.
Experiments across mathematical and general reasoning benchmarks demonstrate consistent improvements, suggesting that temporal scheduling constitutes a promising optimization dimension.
\end{abstract}
\begin{figure*}[!htb]
    \centering
    \includegraphics[width=0.99\linewidth]{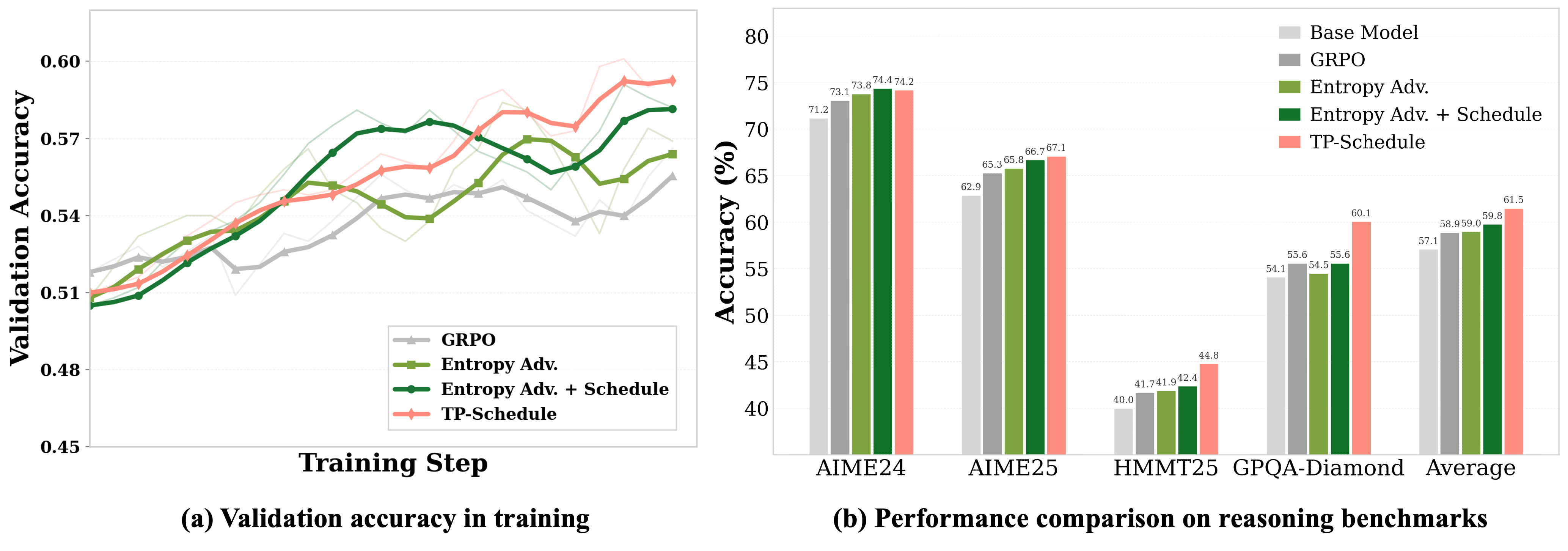}
    \vspace{-1em}
    \caption{Performance and training dynamics on Qwen3-4B model. (a) Temporal scheduling shows stable training dynamics across both existing credit allocation method, e.g., entropy-based advantage reweighting, and the identified trajectory percentiles scheduling, i.e., TP-Schedule, that extends optimization from later to earlier tokens, consistently pushing the optimization plateau to a higher ceiling throughout training. (b) Temporal scheduling consistently outperforms vanilla GRPO and corresponding stagnant credit allocation baselines over representative reasoning benchmarks. }
    \label{fig:val-curve}
    \vspace{-1em}
\end{figure*}
\section{Introduction}
Reinforcement learning with verifiable rewards (RLVR) has emerged as a central paradigm for post-training of Large Language Models (LLMs), yielding substantial improvements in reasoning performance across a wide range of tasks~\citep{guo2025deepseek,comanici2025gemini,yang2025qwen3,team2026kimi}.
Unlike supervised fine-tuning, where dense next-token supervision provides explicit learning signals along the trajectory, RLVR typically broadcasts a single scaler reward across the full sequence, engaging all sampled tokens to participate equally in policy optimization. 
However, the heterogeneous policy behaviors exhibited along the trajectory are largely overlooked, such as reasoning scaffolding and answer convergence, and are optimized roughly without differentiation.

To encourage effective optimization, existing works address this by introducing credit allocation, supposed to identify where the learning signals are distributed within the sampled response.
Representative approaches include Process Reward Models (PRMs) that provide step-level feedbacks~\citep{zou2025reasonflux,zhao2025genprm}, while the heavy annotation cost and computation burden pose a significant scalability bottlenecks.  
More recent works instead focus on token-level signals derived from policy proxies for credit allocation, such as token entropy~\citep{cheng2025reasoning}, likelihood~\citep{damani2025beyond,yang2025not}, and KL divergence~\citep{ma2026fipo,huang2026direction}.
These methods typically identify specific policy behaviors, such as logic connection and reflection, with predefined allocation criteria, and reweight token advantages for preference in policy optimization, resulting in improved performance.
Consistently, parallel studies further demonstrate that optimizing only a subset of critical tokens can outperform full-token optimization~\citep{wang2025beyond}, reinforcing the view that policy gradients accumulated uniformly from all sampled tokens are suboptimal, and could be more effective through more targeted optimization criteria. 

Despite progress, existing credit allocation methods are primarily restricted to stagnant allocation criteria throughout training, where once a token-level proxy is applied, the corresponding targeted policy behavior is continuously emphasized during the entire optimization. 
In view of this, the effectiveness of policy optimization is fundamentally determined by how well the adopted proxy in capturing the informative learning signals, while the behaviors outside the proxy identification could be suppressed for insufficient optimization.
We thereby asking, \textit{instead of persistently pursuing effective proxies for directing policy optimization, are there other potential optimization dimensions that have been underexplored.} 
From the perspective of training dynamics, we advocate that a single credit allocation criteria should not dominate the entire policy optimization process in practice, which is essentially different from their original design choice within individual sample.

Building on this principle, we revisit RLVR optimization from temporal dimension and argue that \textit{when} learning signals are scheduled throughout training can be as important as \textit{where} they are allocated across sampled tokens.
Our key idea is to schedule the allocation criteria over the course of optimization, which prioritizing targeted tokens that emphasized by corresponding allocation criteria, and gradually attenuating toward general optimization. 
We show that temporally scheduling the allocation criteria complements existing credit allocation methods, including both advantage reweighting, and sparse token optimization.
Furthermore, we identify that simple trajectory percentiles serve as an effective criteria for scheduling policy optimization, supported by its natural perspective in distinguishing heterogeneous policy behaviors.
Specifically, we schedule from later tokens for trustworthy continuation, and incrementally incorporate earlier tokens for reliable reasoning scaffolding.
In addition, we reveal that the policy entropy is largely sacrificed in standard RLVR optimization for accommodating heterogeneous policy behaviors, whereas temporal scheduling mitigate this effect with healthier optimization dynamics. 
Experiments across model scales and RLVR algorithms demonstrate consistent improvements, suggesting a promising dimension for RLVR optimization.

The contributions of this work can be summarized as follows: 
(1) We identify the temporal dimension beyond conventional stagnant credit allocation paradigm for RLVR optimization, and propose temporally scheduling the allocation criteria over the course of training, that prioritizing credit allocated tokens, and gradually attenuating toward general optimization.
(2) Temporal scheduling complements existing credit allocation methods, including continuous advantage reweighting, and discrete sparse token optimization. We further identify simple trajectory percentiles as a natural perspective in distinguishing policy behaviors, which works effectively with temporal scheduling.
(3) Experiments across models scales, RL algorithms, and credit allocation strategies show consistent improvement, and further analysis show that temporal scheduling substantially mitigates the sacrificed policy entropy induced by accommodating heterogeneous policy behaviors in RLVR optimization.
\begin{figure*}[t]
    \centering
    \includegraphics[width=0.99\linewidth]{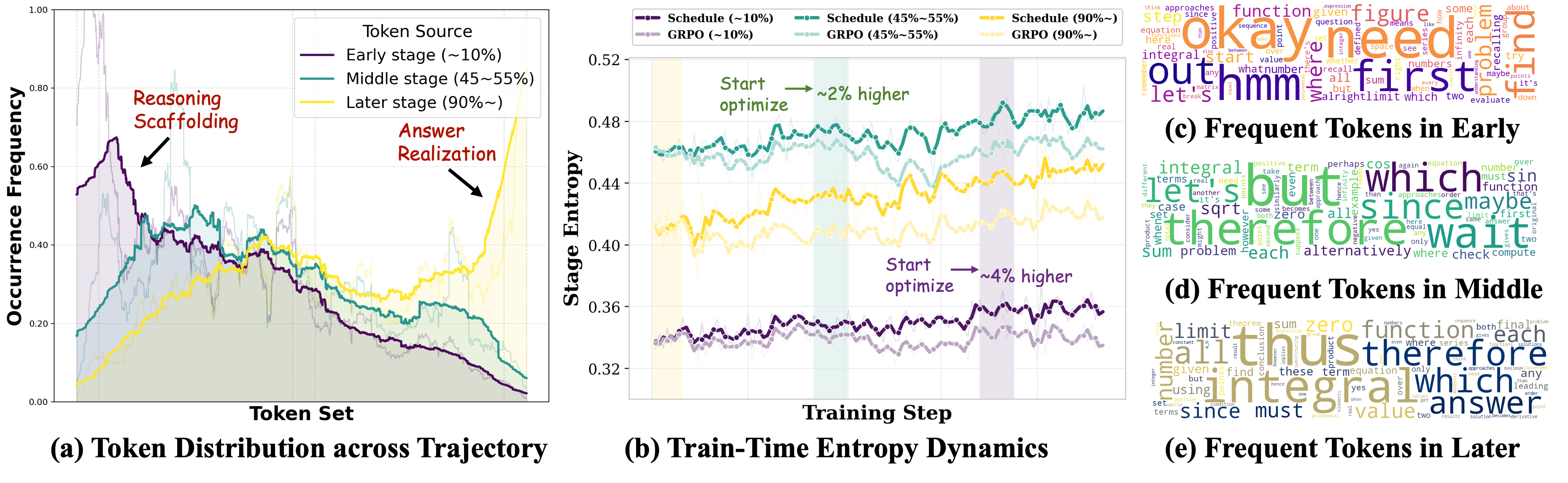}
    \caption{Policy behavior analysis and corresponding training dynamics. (a) Histogram of sampled token frequency across different trajectory positions. The x-axis represents the union of all sampled tokens, sorted by \textit{TP-Score} (Eq.~\ref{eq:tps_score}) to approximate the most likely position of each token along the trajectory. While different trajectory position exhibit substantially different policy behavior with respective token distribution. (b) Entropy dynamics of tokens at different trajectory percentiles. Tokens from different positions exhibit different entropy profiles, while temporal scheduling consistently preserves higher entropy throughout training. (c)-(e) Textual visualization of trajectory positions.}
    \label{fig:stat}
    \vspace{-1em}
\end{figure*}
\section{Preliminaries}
\label{sec:prelim}
Reinforcement learning with verifiable rewards (RLVR) has emerged as a common paradigm for post-training of large language models.
Consider a language model parameterized by $\pi_{\theta}$.
Given a prompt $x$ sampled from data distribution $\mathcal{D}$, the policy generates a response $y = (y_1, \ldots, y_T)$ of length $T$.
A verifier then assigns a global scalar reward $r(x, y) \in \mathbb{R}$ that measures the quality of the sampled response.
The RLVR objective is to maximize the expected reward
\begin{equation}
J(\theta) = \mathbb{E}_{x \sim \mathcal{D}, \, y \sim \pi_{\theta}(\cdot \mid x)} \left[r(x, y)\right].
\end{equation}

\textbf{Group Relative Policy Optimization (GRPO).} GRPO~\citep{shao2024deepseekmath} simplfies the Proximal Policy Optimization (PPO)~\citep{schulman2017proximal} by eliminating the critic model for advantage estimation.
For each prompt, GRPO samples a group of response $\{\textbf{y}_{i}\}_{i=1}^{G}$ and estimates the advantage of each response in a group-relative manner, and the clipped surrogate objective is preserved as PPO:
\begin{align}
\label{eq:grpo}
\mathcal{J}_\text{GRPO}(\theta) = \mathbb{E}_{x \sim \mathcal{D},\, \scalebox{0.65}{$\{\bm{y}_i\}_{i=1}^G$} \sim \scalebox{0.65}{$\pi_{\theta}( \cdot | x)$}} \left[ \frac{1}{G} \sum_{i=1}^{G} \frac{1}{|y_i|} \sum_{t=1}^{|y_i|} \min \left( w_{i,t}(\theta) \widehat{A}_{i}, \, \mathrm{clip} \left( w_{i,t}(\theta), 1 - {\varepsilon}, 1 + {\varepsilon}\right) \widehat{A}_{i} \right) \right], \nonumber
\end{align}
where $w_{i,t}(\theta)=\frac{ \pi_{\theta} (y_{i,t} | x, y_{i,<t}) }{ \pi_{\theta_\text{old}} (x_{i,t} | x,y_{i,<t})}$ is the importance sampling ratio between the current and old policy models, and is typically approximate 1 under on-policy optimization. The advantage $\widehat{A}_{i}$ is shared across all tokens within response $\bm{y}_i$, and the effective gradient of GRPO objective is derived as:
\begin{align}
\nabla_{\theta} \mathcal{J}_\text{GRPO}& (\theta) 
=\ 
\nabla_{\theta} \mathbb{E}_{ x \sim \mathcal{D},\, \{y_i\}_{i=1}^G \sim \pi_{\theta_\text{old}}( \cdot | x) }
\left[ \frac{1}{G} \sum_{i=1}^{G} \frac{1}{|y_i|} \sum_{t=1}^{|y_i|} 
w_{i,t}(\theta) \widehat{A}_{i,t}
\right] \\
=&\
\mathbb{E}_{ x \sim \mathcal{D},\, \{y_i\}_{i=1}^G \sim \pi_{\theta_\text{old}}( \cdot | x) }
\left[ \frac{1}{G} \sum_{i=1}^{G} \widehat{A}_{i} 
\cdot \frac{1}{|y_i|} \sum_{t=1}^{|y_i|} 
\frac{ \pi_{\theta} (y_{i,t} | x, y_{i,<t}) }{ \pi_{\theta_\text{old}} (y_{i,t} | x,y_{i,<t})} 
\nabla_{\theta} \log \pi_{\theta} (y_{i,t} | x, y_{i,<t})  
\right],
\end{align}
which is typically accumulated as token-level average in driving policy optimization. 
While the weighted importance sampling ratio $w_{i,t}(\theta)$ primarily serves as a distributional correction term in off-policy setting, rather than indicating respective token-level importance.

\textbf{Token-Level Policy Optimization.} 
To encourage efficient policy optimization, recent methods focus on allocating globally distributed advantage $\widehat{A}$ over tokens for emphasizing token importance, as
\begin{align}
\label{eq:credit-gradient}
\nabla_{\theta} \mathcal{J}_\text{GRPO-credit}& (\theta) 
= \mathbb{E}_{ x \sim \mathcal{D},\, \{y_i\}_{i=1}^G \sim \pi_{\theta_\text{old}}( \cdot | x) }
\left[ \frac{1}{G} \sum_{i=1}^{G}
\widehat{A}_{i} \cdot \frac{1}{|y_i|} \sum_{t=1}^{|y_i|} 
\psi(\mu_{i,t})
g_{i,t}(\theta)
\right],
\end{align}
where the $g_{i,t}(\theta)= w_{i,t}(\theta) \cdot \nabla_{\theta} \log \pi_{\theta} (y_{i,t} | x, y_{i,<t}) $ denotes the per token policy gradient, $\mu_{i,t}$ denotes the derived policy proxy for token $y_{i,t}$, and $\psi$ is a wrap function for proxy utilization.
While the allocated $\widehat{A}_{i} \cdot \psi(\mu_{i,t})$ effectively determine the magnitude of each token in contributing to policy optimization. We mainly identifying two types of credit allocation paradigms as following:
\begin{itemize}[leftmargin=2em]
\item Advantage Reweighting: The credit allocation factor $\mu_{i,t}$ can derive from diverse policy-derived proxies, such as token entropy~\citep{cheng2025reasoning} and $\psi$ typically denotes continuous function.
\item Sparse Optimization: ~\citep{wang2025beyond} only optimize a subset of critical tokens that selected by $\mu_{i,t}$, and $\psi$ typically formed as indicator function for deriving binary mask.
\end{itemize}

\textbf{Trajectory Percentile Score.}
\label{sec:pre_tps}
To investigate whether the policy $\pi_{\theta}$ emerges distinct behaviors along the trajectory, we introduce the Trajectory Percentile Score (\textit{TP-Score}), that characterizes the most likely position the token is sampled along the trajectory.
Consider a set of predefined position ranges $\mathcal{P}=\{(p_i-\delta,p_i+\delta)\}_{i=1}^{N}$ along the trajectory, where $p_i \in (0,1]$ denotes the center of $i$-th range and normalized by trajectory length, $\delta$ denotes the radius.
We record the occurrence frequency of each sampled tokens $y$ within each position ranges, denoted as $c_i(y)$, and the \textit{TP-Score} is as
\begin{align}
\label{eq:tps_score}
\mathbb{E}_{i\sim \tilde{c}_i(y)}[p_i]= \frac{\sum_{i} c_i(y) \cdot p_i}{\sum_{i} c_i(y)},  
\end{align}
where $\tilde c_i(y)=\frac{c_i(y)}{\sum_j c_j(y)}$ denotes the approximate probability of token $y$ sampled at $i$-th position range, and smaller \textit{TP-Score} correspond to earlier positions and larger values correspond to later.
\section{Method}
\label{sec:method}
We first analyze the distribution of tokens sampled at different positions, showing that trajectory percentile serves as a natural perspective in distinguishing policy behaviors and revealing distinct training dynamics. 
To explore effective RLVR optimization, we introduce temporal dimension that schedules token optimization over the course of training, complementing existing stagnant credit allocation methods, including advantages reweighting and sparse optimization.
Finally, we show that simple trajectory percentiles can effectively integrated with temporal scheduling.

\subsection{Demystifying Policy Behavior Through Trajectory Percentile}
\label{sec:analysis}
Previous studies have shown that the policy exhibits distinct behaviors across sampled tokens, such as forking tokens for logical connection and following tokens for sentence completion~\citep{wang2025beyond}, which have been effectively demystified through token entropy, and broadly adopted in subsequent works as proxies for credit allocation~\citep{cheng2025reasoning}.
In this work, we examine whether policy behaviors can be naturally distinguished along the trajectory.
We use Qwen3-4B~\citep{yang2025qwen3} to generate rollouts on a challenging 78k subset curated from OpenMathReasoning~\citep{moshkov2025aimo2} and DeepMath-103K~\citep{he2025deepmath}, which have been processed through deduplication, resulting in 584M response tokens in total.
We then introduce three representative trajectory position ranges, i.e., $\{p_i\}_{i=1}^N=\{0.05, 0.5, 0.95\}$ as described in Sec.~\ref{sec:pre_tps}, 
and the range radius $\delta$ is set by $0.05$. 
We show the sampled token distribution at different position ranges in Figure~\ref{fig:stat} (a), where the x-axis represents the union of all sampled tokens within $\mathcal{P}$, sorted by the \textit{TP-Score} to approximate the trajectory position where the tokens are most likely to be sampled, as detailed in Sec.~\ref{sec:pre_tps}.

\textbf{Trajectory percentiles provide a natural perspective in distinguishing policy behavior.} 
We observe substantial differences in sampled token distributions across trajectory percentiles, where the tokens sampled from early percentiles, i.e., $p_i=0.05$, predominantly occur near the beginning of the trajectory, and rarely appear at later positions. 
While tokens sampled from middle and later percentiles exhibit similar predominant sampling behavior around the corresponding positions, suggesting that heterogeneous policy behaviors can be naturally distinguished along the trajectory.

\textbf{Trajectory percentile anchors the policy entropy in orchestrating optimization.} 
In Figure~\ref{fig:stat} (b), we show that the policy entropy diverge substantially across different trajectory percentiles, both at initialization and throughout training, suggesting that trajectory percentiles provide an effective anchor of policy entropy dynamics for accommodating optimization.
In particular, early trajectory percentiles exhibit the lowest entropy, while the middle percentiles consistently maintain the highest entropy.
We provide textual visualizations in Figure~\ref{fig:stat} (c)-(e), where distinct policy behaviors can be observed across trajectory.
Early percentiles primarily correspond to reasoning scaffolding behavior, and tend to follow constrained patterns in the absence of sufficient context. 
While the middle percentiles are more associated with reflection and critical lighting behavior,exhibiting higher exploratory characteristics under intermediate context states. 
Later percentiles are mainly responsible for answer realization and summarization, exhibiting entropy falling in between.

\begin{figure}[t]
  \centering
  \begin{algorithm}[H]
    \caption{Temporal Scheduling for Credit Allocation}
     \label{alg:online_deepconf}
   \begin{algorithmic}
     \STATE {\bfseries Inputs:} Train data $\mathcal{D}\{x,y\}$, policy $\pi_{\theta}$, total training steps $N$, credit allocation criteria $\psi(\mu_{i})$, token proxy $\mu_{i}$, allocation function $\psi(\cdot)$, schedule function $\mathcal{S}(\cdot)$, temporal range $\varepsilon_{low}$, $\varepsilon_{high}$.
     \STATE {\bfseries Outputs:} Optimized policy $\pi$.
     \STATE Initialize training progress $\tau \leftarrow 0$ .
     \WHILE{$\tau \leq N$}
     \STATE Determine the scheduled progress $\bm{\tau}$ from Eq.~\ref{eq:schedule}.
     \STATE Generate rollouts for batch data from $\mathcal{D}$, and estimate the credit allocation factor for each sampled token by $\psi(\mu_{t})$.
     \STATE Schedule the allocated $\psi(\mu_t)$ with Eq.~\ref{eq:continuous} and Eq.~\ref{eq:discrete}.
     \STATE Optimize policy $\pi_{\theta}$ on batch data with scheduled advantage using Eq.~\ref{eq:credit-schedule}.
     \STATE $\tau = \tau + 1$.
     \ENDWHILE
     \RETURN policy $\pi_{\theta}$, and end training.
   \end{algorithmic}
  \end{algorithm}
  \vspace{-2.4em}
\end{figure}
\subsection{Temporal Scheduling of Credit Allocation}
\label{sec:method:schedule}
RLVR optimizes policies by broadcasting a single scaler reward over the full sequence, yielding reliable directional supervision, while suffer from sparse magnitude signals along the trajectory~\citep{yang2026self}.
Existing works address this by introducing credit allocation strategies, typically using policy-derived proxies, such as token entropy, to reweight token-level advantages and reinforce specific policy behaviors, such as logic connection and reflection. 
While the allocation criteria are principally stagnant throughout training, limiting resilient policy evolution and performance ceiling, as illustrated in Figure~\ref{fig:val-curve}.
Building upon this, we expose that temporally scheduling the allocation criteria over the course of training provides a promising optimization dimension, where the scheduled credit allocation is formed as

\begin{equation}
\label{eq:schedule}
    \widehat{f}_{t(\tau)} = \mathcal{S}(\bm{\tau})\widehat{f}_{t}, \quad \text{where} \; \bm{\tau} = \frac{\mathrm{clip}(\tau,\, \tau_{low},\, \tau_{high})-\tau_{low}}{\tau_{high}-\tau_{low}}.
\end{equation}
We denote $\widehat{f}_{t}$ as the original credit allocation factor over token $y_t$, yielded from existing methods.
$\tau$ denotes the current training steps, and $\tau_{high}$ and $\tau_{low}$ determine the temporal range over which scheduling is applied.
$\mathcal{S}(\bm{\tau}) \in [0,1)$ is a monotonically decreasing schedule function.
We show that the scheduled credit $\widehat{f}_{t(\tau)}$ starts from the most targeted allocation criteria, i.e., $\widehat{f}_{t}$, to reinforce specific policy behaviors, and gradually attenuates toward general optimization over the course of training.

\textbf{Integrating into Policy Optimization.} The temporal scheduled credit $\widehat{f}_{t(\tau)}$ serves as a drop-in replacement for original allocation factor $\widehat{f}_{t}$ in policy optimization, as referenced in Eq.~\ref{eq:credit-gradient}. 
We demonstrate its applicability to two representative allocation paradigms that define $\widehat{f}_{t}$, including \textit{advantage reweighting} and \textit{sparse optimization}.
As described in Sec.~\ref{sec:prelim}, we formulate $\widehat{f}_{t}=\psi(\mu_{t})$, where the policy proxy $\mu_t$ (e.g., token entropy) is wrapped by function $\psi$ for credit utilization.
In view of this, the \textit{advantage reweighting} paradigm typically form $\psi$ as a continuous function, yielding positive scaler without altering the sign of the advantage.
While \textit{sparse optimization} defines $\psi$ as indicator functions $\mathbb{I}\big(\mu_{t} \ge \varepsilon\big)$ with threshold $\varepsilon$, yielding binary masks that select a subset of tokens for optimization.
We show that integrate the temporal scheduling on such cases extends Eq.~\ref{eq:schedule} as 

\begin{align}
\mathrm{Continuous:}\quad
& \widehat{f}_{t{(\tau)}}
= \mathcal{S}(\bm{\tau})\psi(\mu_{t}), \quad \text{where} \; \widehat{f}_t \ge 0, \label{eq:continuous}\\
\mathrm{Discrete:}\quad 
& \widehat{f}_{t(\tau)}
= \mathbb{I}\big(\mu_{t} \ge \mathcal{S}(\bm{\tau}) \cdot\varepsilon\big),\quad \text{where} \; \widehat{f}_t \in \{0,1\}. \label{eq:discrete}
\end{align}
The continuous case follows the same as Eq.~\ref{eq:schedule}, where the derived allocation factor $\psi(\mu_{t})$ is temporally scheduled throughout training. 
While for discrete case, we instead schedule the threshold $\varepsilon$ over the course of training, gradually relaxing the criterion and including more tokens for optimization.

\subsection{Scheduling over Trajectory Percentiles}
Instead of deriving the proxy $u_t$ from prevalent policy measures such as entropy to identify informative learning signals, we show that temporal scheduling works more effectively with simple trajectory percentiles.
As analyzed in Sec~\ref{sec:analysis}, different trajectory percentiles exhibit substantially distinct sampled token distribution, providing a natural perspective in distinguishing heterogeneous policy behaviors, and evolving optimization throughout training.
In particular, we prioritize later-percentile tokens that are closer to final answer realization, and gradually expand optimization towards earlier tokens that responsible for reasoning scaffolding. 
The percentile schedule over trajectory follows the discrete allocation scheduling in Eq.~\ref{eq:discrete}, and is formed as
\begin{equation}
\widehat{f}_{t(\tau)} = \mathbb{I}\left[\mu_t \geq \mathcal{S}(\bm{\tau}) \cdot1\right], \quad \text{where} \; \mu_t = \frac{t}{T} \in [0, 1),
\end{equation}
where $\mu_t$ is defined as trajectory percentile of token $y_t$, normalized by the response length $T$ of trajectory $y$.
$\varepsilon=1$ denotes the maximal percentiles at initialization, and $\mathcal{S}(\bm{\tau}) \in [0,1)$ monotonically decrease the percentile criterion to include earlier tokens into optimization.
In the case of GRPO, the effective policy gradients under temporal scheduling over trajectory percentile is given by
\begin{align}
\label{eq:credit-schedule}
\nabla_{\theta} \mathcal{J}_\text{GRPO-schedule}& (\theta) 
= \mathbb{E}_{ x \sim \mathcal{D},\, \{y_i\}_{i=1}^G \sim \pi_{\theta_\text{old}}( \cdot | x) }
\left[ \frac{1}{G} \sum_{i=1}^{G}
\widehat{A}_{i} \cdot \frac{1}{|y_i|} \sum_{t=1}^{|y_i|} 
\mathbb{I}\left[\frac{t}{|y_i|} \geq \mathcal{S}(\bm{\tau}) \cdot 1\right]
g_{i,t}(\theta)
\right],
\end{align}
where $g_{i,t}(\theta)$ denotes the token-level policy gradient, as described in Eq.~\ref{eq:credit-gradient}.
The proposed trajectory percentile scheduling manages policy improvements by accumulating policy gradients over targeted tokens and evolves optimization along trajectory, which initially focus on reliable downstream continuations and gradually calibrates earlier behaviors once the policy has been partially stabilized.
\section{Experiments}
\label{sec:experimenal}
\subsection{Experimental Setup}
\textbf{Train Data and Benchmarks.}
We train our models on a challenging 30K dataset curated from OpenMathReasoning~\citep{moshkov2025aimo2} and DeepMath-103K~\citep{he2025deepmath}, and processed through deduplication and difficulty filtering.
For each problem, we use DeepSeek-V3.2-Speciale~\citep{liu2025deepseek} to synthesize three reasoning trajectories, and only those correctly verified at least one time are retained.
We evaluate on in-distributional mathematical benchmarks: AIME24/25~\footnote{https://huggingface.co/datasets/AI-MO/aimo-validation-aime}, HMMT25~\citep{balunovic2025matharena}, Minerva~\citep{lewkowycz2022solving}, and OlympiadBench~\citep{he2024olympiadbench}, and out-of-distributional reasoning benchmarks, including Winogrande~\citep{sakaguchi2021winogrande} for commonsense reasoning, GPQA-Diamond~\citep{rein2023gpqa} for STEM reasoning, and MuSR~\citep{sprague2023musr} for multistep soft reasoning.
We use a rollout temperature of 0.6, and top-\textit{p} sampling with \textit{p}=0.95, under a maximum response length of 32768.
We report the Avg@16 metric for AIME24/25 and HMMT25, and report the Avg@4 metric for other benchmarks.

\begin{table*}[t]
\caption{Performance across mathematical and general reasoning benchmarks. \textbf{Bold} indicates the best results among the comparison. `` $\hookrightarrow$Schedule '' represents applying the temporal schedule to corresponding credit allocation methods, while `` TP '' denotes the trajectory percentile.}
\vspace{0.2em}
\label{tab:main_results}
\centering
\small
\setlength{\tabcolsep}{0.002pt}
\renewcommand{\arraystretch}{1.18}
\begin{tabular}{lcccccccccc}
\toprule
\multirow{2}{*}{\textbf{Model}}
& \multicolumn{5}{c}{\textbf{In-Distribution}}
& \multicolumn{4}{c}{\textbf{Out-of-Distribution}} \\
\cmidrule(lr){2-6} \cmidrule(lr){7-10}
& AIME24
& AIME25
& HMMT25
& Minerva
& \textbf{Avg.}
& WinoG*
& GPQA*
& MuSR
& \textbf{Avg.}\\
\midrule

Qwen3-4B
& 71.2 & 62.9 & 40.0 & 79.8 &63.5
& 77.7 & 54.1 & 61.4 & 64.4 \\
GRPO
& 73.1 & 65.3 & 41.7 & 80.9 & 65.3
& 76.8 & 55.6 & 60.9 & 64.4\\
\rowcolor[HTML]{F4F6FB} TP-Schedule
& \hspace{2.8mm}\gainvalue{\textbf{74.2}}{1.1}{1.3} & \hspace{2.8mm}\gainvalue{\textbf{67.1}}{1.8}{1.3} & \hspace{2.8mm}\gainvalue{\textbf{44.8}}{3.1}{1.3} & \hspace{2.8mm}\gainvalue{\textbf{83.8}}{2.9}{1.3} & \hspace{2.8mm}\gainvalue{\textbf{67.5}}{2.2}{1.3}& 
\hspace{2.8mm}\gainvalue{\textbf{78.4}}{1.6}{1.3} & \hspace{2.8mm}\gainvalue{\textbf{60.1}}{4.5}{1.3} & \hspace{2.8mm}\gainvalue{\textbf{62.9}}{2.0}{1.3} & \hspace{2.8mm}\gainvalue{\textbf{67.1}}{2.7}{1.3} \\
GSPO
& 71.5 & 65.6 & 43.3 & 79.8 & 65.0
& 76.6 & 54.5 & 61.5 & 64.2\\
\rowcolor[HTML]{F4F6FB} TP-Schedule
& \hspace{2.8mm}\gainvalue{72.3}{0.8}{1.3} & \hspace{2.8mm}\gainvalue{66.8}{1.2}{1.3} & \hspace{2.8mm}\gainvalue{44.0}{0.7}{1.3} & \hspace{2.8mm}\gainvalue{80.2}{0.4}{1.3} & \hspace{2.8mm}\gainvalue{65.8}{0.8}{1.3}& 
\hspace{2.8mm}\gainvalue{78.1}{1.5}{1.3} & \hspace{2.8mm}\gainvalue{57.1}{2.6}{1.3} & 
\hspace{2.8mm}\gainvalue{62.4}{0.9}{1.3} &
\hspace{2.8mm}\gainvalue{65.9}{1.7}{1.3} \\

\midrule
Entropy Adv.
& 73.8 & 65.8 & 41.9 & 80.1 & 65.4
& 77.8 & 54.5 & 61.7 & 64.6\\
\rowcolor[HTML]{F4F6FB}$\hookrightarrow$ Schedule
& \hspace{2.8mm}\gainvalue{\textbf{74.4}}{0.6}{1.3} & \hspace{2.8mm}\gainvalue{\textbf{66.7}}{0.9}{1.3} & \hspace{2.8mm}\gainvalue{42.4}{0.5}{1.3} & 
\hspace{2.8mm}\nonevalue{80.1}{0.0}{1.3} & 
\hspace{2.8mm}\gainvalue{65.4}{0.8}{1.3}
& \hspace{2.8mm}\gainvalue{\textbf{78.5}}{0.7}{1.3} & 
\hspace{2.8mm}\gainvalue{55.6}{1.1}{1.3} & 
\hspace{2.8mm}\gainvalue{62.2}{0.5}{1.3} & 
\hspace{2.8mm}\gainvalue{65.4}{0.8}{1.3} \\
Forking Tok.
& 71.5 & 64.8 & 41.5 & 81.3 & 64.8
& 77.6 & 55.6 & 61.5 & 64.9\\
\rowcolor[HTML]{F4F6FB}$\hookrightarrow$ Schedule
& \hspace{2.8mm}\gainvalue{74.2}{2.7}{1.3} & \hspace{2.8mm}\gainvalue{65.8}{1.0}{1.3} & \hspace{2.8mm}\gainvalue{\textbf{42.7}}{1.2}{1.3} & \hspace{2.8mm}\gainvalue{\textbf{82.1}}{0.9}{1.3} & \hspace{2.8mm}\gainvalue{\textbf{66.2}}{1.4}{1.3} & 
\hspace{2.8mm}\gainvalue{78.2}{0.6}{1.3} & \hspace{2.8mm}\gainvalue{\textbf{56.3}}{0.7}{1.3} & \hspace{2.8mm}\gainvalue{\textbf{63.2}}{1.7}{1.3} & 
\hspace{2.8mm}\gainvalue{\textbf{65.9}}{1.0}{1.3} \\

\midrule
Qwen3-8B
& 74.8 & 66.3 & 41.8 & 82.3 & 66.3
& 81.3 & 59.6 & 62.6 & 67.9  \\
GRPO
& 75.6 & 68.6 & 42.7 & 83.6 & 68.0
& 82.0 & 62.7 & 63.9 & 69.5 \\
\rowcolor[HTML]{F4F6FB}TP-Schedule
& \hspace{2.8mm}\gainvalue{\textbf{76.4}}{0.8}{1.3} & \hspace{2.8mm}\gainvalue{\textbf{70.0}}{1.4}{1.3} & \hspace{2.8mm}\gainvalue{\textbf{44.3}}{1.6}{1.3} & \hspace{2.8mm}\gainvalue{\textbf{84.5}}{0.9}{1.3} & \hspace{2.8mm}\gainvalue{\textbf{68.8}}{0.8}{1.3}
& \hspace{2.8mm}\gainvalue{\textbf{82.8}}{0.8}{1.3} & \hspace{2.8mm}\gainvalue{\textbf{64.1}}{1.4}{1.3} & \hspace{2.8mm}\gainvalue{\textbf{64.5}}{0.6}{1.3} &
\hspace{2.8mm}\gainvalue{\textbf{70.5}}{1.0}{1.3} \\
\bottomrule
\end{tabular}
\vspace{-1.5em}
\end{table*}

\textbf{Models and Baselines.}
We perform our main experiments on Qwen3-4B and Qwen3-8B~\citep{yang2025qwen3}.
We compare our method primarily against vanilla GRPO~\citep{shao2024deepseekmath}, and advanced token-level optimization baselines, including Entropy-Adv.~\citep{cheng2025reasoning} that performs entropy-based advantage reweighting, and Forking-Tok.~\citep{wang2025beyond} that only optimizes a subset of forking tokens, selected by entropy proxy for policy gradients accumulation.
Temporally scheduling the corresponding continuous and discrete credit allocation criteria are performed for comparison.
TP-Schedule represents the trajectory percentile schedule on basic RL algorithms without other advanced credit allocation.
We additionally validate our schedule on different RLVR algorithms including GSPO~\citep{zheng2025group}, which underscores sequence-level optimization.

\textbf{Implementation Details.}
We implement our methods based on the VERL~\citep{sheng2024hybridflow} framework.
The maximum model length is set to 40960 tokens, with a maximum prompt length of 1024 and a maximum response length of 39936 to fully utilize the context window.
We use the learning rate of $1\times10^{-6}$ and the training batch size of 256.
For each prompt, we sample 8 rollouts with a temperature of 1.2.
Clipping thresholds are set to $\epsilon_{\mathrm{low}}=0.2$ and $\epsilon_{\mathrm{high}}=0.28$, and we exclude both KL penalty and entropy regularization.
We adopt the linear schedule as $\mathcal{S}(\cdot)$ in our main experiments, and apply temporal range $\tau_{low}=0$ and $\tau_{high}=0.8$, normalized by total training steps.
Unless otherwise noted, we use the recommended implementation configuration and keep the training data, and evaluation settings consistent across methods for fair comparison.

\subsection{Main Results}
Table~\ref{tab:main_results} presents the evaluation results on in-distribution mathematical benchmarks and out-of-distribution multi-domain reasoning benchmarks, where the temporal scheduling show consistent improvement on both existing credit allocation methods, and proposed trajectory percentiles criteria.
(1) On Qwen3-4B model, scheduling with trajectory percentiles over the course of training, i.e., TP-Schedule, outperforms the vanilla GRPO by 2.2\% on mathematical and 2.7\% on general reasoning benchmarks, respectively.
While the improvements on Qwen3-8B further indicate that the trajectory percentile serves as a simple but effective criteria in working with temporal schedule, with overall 1\% improvements over vanilla GRPO.
(2) Additionally, for existing stagnant credit allocation methods, scheduling the corresponding allocation criteria, such as continuous advantage reweighting factor in Entropy Adv., and discrete threshold of selecting optimization tokens in Forking Tok., both show that temporal dimension provides a promising optimization direction for policy evolution.  
We observe that discretely expanding the number of tokens involved in optimization show more potential, suggesting that the accumulated policy gradients from gated tokens could be more efficient than soft weighting across all tokens.
(3) We also validate the temporal scheduling on other RLVR algorithms, such as GSPO, where the performance show consistent improvement.

\begin{figure*}[!htb]
    \centering
    \includegraphics[width=0.99\linewidth]{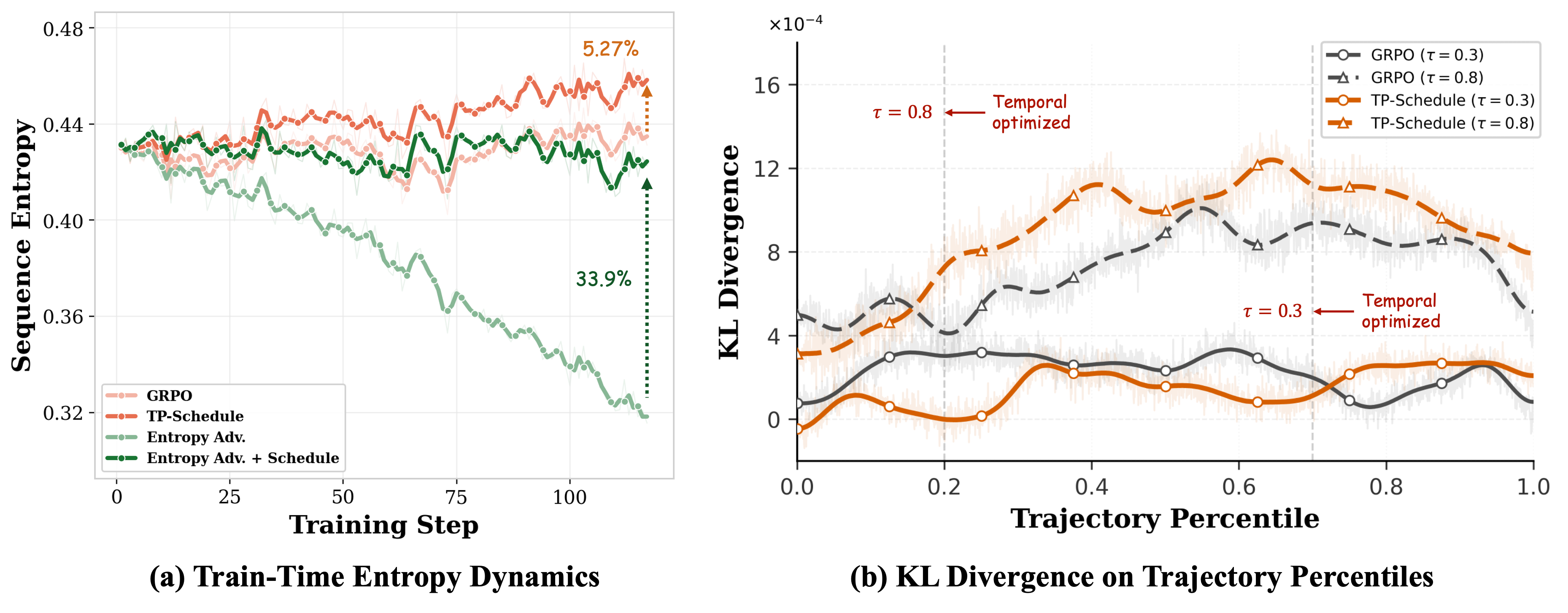}
    \vspace{-0.8em}
    \caption{Analysis of policy optimization under temporal scheduling. (a) Temporal scheduling consistently preserves higher policy entropy over full response under different credit allocation strategies. (b) Temporal scheduling over trajectory induces a structured shift in KL divergence from later to earlier percentiles over training, suggesting the non-uniformly optimized policy behaviors.}
    \label{fig:exp-curve}
    \vspace{-1.5em}
\end{figure*}
\subsection{Temporal Scheduling Analysis}
Beyond aggregate performance gains, we observe several characteristic phenomena that shed light on why temporal scheduling is effective in RLVR. 
By jointly analyzing training dynamics and trajectory-level behaviors, we identify two key factors that drive its improvements: 
the consistently reserved entropy compared to standard baselines, and the progressive alignment of policy behaviors.

\textbf{Entropy preservation under temporal scheduling.}
We first examine the training-time entropy dynamics in Figure~\ref{fig:exp-curve} (a), where the temporal scheduling and corresponding baselines are evaluated under different credit allocation strategies.
Compared to standard GRPO, which optimized with all sampled tokens, temporal scheduling along trajectory percentiles consistently preserves higher policy entropy on full sequence. 
This effect is particularly pronounced when compared with entropy-based advantage reweighting, where entropy collapses rapidly as optimization aggressively concentrates on high entropy tokens, while temporally scheduling results more steady optimization and preserves policy entropy over 33.9\%, and accompanied with higher performance in Table~\ref{tab:main_results}.

\textbf{Temporal alignment of policy behaviors along the trajectory.} 
We further analyze the token-level KL divergence normalized across trajectory percentiles in Figure~\ref{fig:exp-curve} (b), which reveals how policy updates are distributed along the trajectory. 
We compute the KL divergence between the model checkpoints at $\bm{\tau}=0.3$ and $\bm{\tau}=0.8$ of the scheduled progress in training with the initial model, where temporal scheduling induces a structured and progressive shift in KL divergence.
Early in training, larger deviations are concentrated on later trajectory percentiles compared to standard GRPO, and as training proceeds, the deviation gradually propagates toward earlier percentiles, suggesting that the policy behaviors are not updated uniformly, but instead optimized temporally for accommodation.

\subsection{Ablations}
To validate the reliability of temporal scheduling, we conduct ablations on Qwen3-4B model, including the sensitivity of schedule choice $\mathcal{S}(\cdot)$, the recommended temporal range to apply scheduling, and the allocation criteria worked with scheduling.

\textbf{The choice of schedule $\mathcal{S}(\cdot)$ has minor effect on performance.} 
We present the results of different scheduling functions applied to attenuate the credit allocation criteria in Table~\ref{tab:abl-schedule}, and report Avg@16
\begin{wraptable}{r}{0.48\textwidth}
\centering
\vspace*{-6.6mm}
\caption{Ablations results of schedule choice.} 
\vspace*{0.8mm}
\resizebox{\linewidth}{!}{
\begin{tabular}{l|ccc}
\toprule
$\mathcal{S}(\cdot)$  & AIME25&  HMMT25 & GPQA-d\\
\midrule
$\varnothing$ &\textcolor[gray]{0.6}{65.3} &  \textcolor[gray]{0.6}{41.7} & \textcolor[gray]{0.6}{55.6}  \\
Sigmoid &66.7 &  43.5 & 58.5  \\
Gamma &66.4 & 43.7  & 57.5 \\
\rowcolor[HTML]{F4F6FB}Linear &67.1 & 44.8  & 60.1\\
\bottomrule
\end{tabular}
}
\label{tab:abl-schedule}
\vspace*{-5mm}
\end{wraptable}
performance on AIME25 and HMMT25, and Avg@4 performance on GPQA-Diamond, where the trajectory percentile is adopted as default allocation criteria.
As described in Appendix~\ref{sec:schedule_function}, the sigmoid features more optimization steps at the beginning and end of scheduling, 
while the gamma stresses more on early stage.
We show that the temporal scheduling is insensitive to specific schedule choice, and show consistent improvement over basic GRPO, denoted as $\varnothing$.
We adopt the linear schedule as the default setting. 

\textbf{Temporal scheduling is more effective when extended to the later stages of training.} 
In Table~\ref{tab:abl-range}, we show results of different temporal range applied to scheduling the policy optimization.
\newcolumntype{C}{>{\columncolor[HTML]{F4F6FB}}c}
\begin{wraptable}{r}{0.52\textwidth}
\renewcommand{\arraystretch}{1.12}
\centering
\vspace*{-4.9mm}
\caption{Ablations results of applied temporal range.} 
\vspace*{0.8mm}
\resizebox{\linewidth}{!}{
\begin{tabular}{l|cccCCc}
\toprule
$\varepsilon_{high}$  & 0.0 & 0.2&  0.4 & 0.6 & 0.8 & 0.95\\
\midrule
AIME25 &\textcolor[gray]{0.6}{65.3}&65.4& 65.4& 66.7& 67.1& 65.6 \\
HMMT25 &\textcolor[gray]{0.6}{41.7}&39.8& 40.8& 43.3& 44.8& 42.6\\
GPQA* &\textcolor[gray]{0.6}{55.6}&53.5& 54.1& 56.6& 60.1& 55.5\\
\bottomrule
\end{tabular}
}
\label{tab:abl-range}
\vspace*{-5mm}
\end{wraptable}
We remark that the temporal scheduling allocates the strongest credit at initial, resulting in $\varepsilon_{low}=0$, where only $\varepsilon_{high}$ is ablated to determine the overall attenuation range.
We notice that extending scheduling to the later stages of training reveals the benefits of the temporal dimension, which allows more sufficient orchestration of policy optimization for accommodate distinct policy behaviors.
We recommend that $\varepsilon \in [0.6,0.8]$ represents a sweet spot for performance, where the allocation criteria over the course of training are well balanced, enabling effective policy optimization.

\begin{wrapfigure}{r}{0.49\textwidth}
    \centering
    \vspace{-0.6em}
    \includegraphics[width=\linewidth]{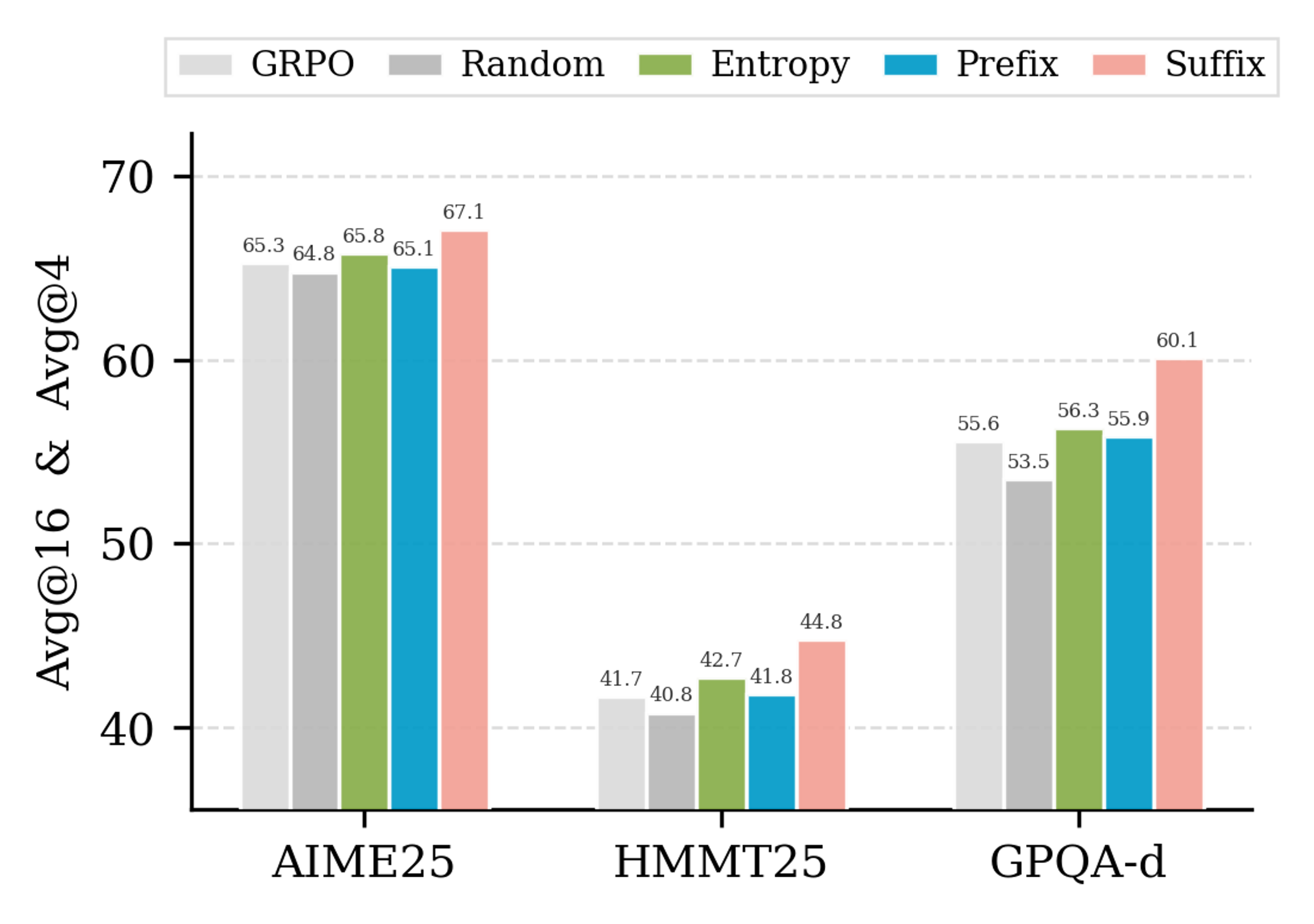}
    \vspace{-2.3em}
    \caption{
    Results of temporal scheduling across different credit allocation proxies.
    }
    \label{fig:ablation_proxy}
    \vspace{-1em}
\end{wrapfigure}

\textbf{Trajectory percentile works effectively with temporal scheduling.}
In Figure~\ref{fig:ablation_proxy}, we show that temporal scheduling exhibits improved performance on reasonable credit allocation proxies, such as entropy
and suffix, where the policy behaviors can be effectively accommodated for optimization,  while degrading performance on others.
In particular, randomly selecting a subset of tokens for scheduling policy optimization significantly corrupts the policy gradients, where the underlying policy behaviors are disorganized for temporal evolution.
Additionally, trajectory percentile works effectively with temporal scheduling, supported by its natural perspective in distinguishing policy behaviors.
However, the temporal order matters, while scheduling the optimization from earlier to later tokens (i.e., Prefix), critically depends on unreliable continuation, resulting in unstable optimization.
The entropy proxy shows stable improvements when optimized from high entropy tokens, and gradually extends toward all sampled tokens, exhibiting improved performance compared to its stagnant counterpart (i.e., Forking Tok.).
Detailed definitions are provided in Appendix~\ref{sec:allocation_proxy}.
\section{Related Work}

\textbf{Credit Assignment.}
Reinforcement learning with verification rewards faces a fundamental credit assignment challenge, where a single scalar reward is broadcast across long sequences, making it difficult to attribute advantages to specific actions.
Common approach introduces dense signals for credit estimation, such as process reward models (PRMs) for step-level feedbacks~\citep{zou2025reasonflux}, and policy proxies including entropy, future KL~\citep{cheng2025reasoning,ma2026fipo}, etc., for token-level advantage shaping. 
And more recent works introduce on-policy distillation (OPD)~\citep{lu2025onpolicydistillation} and self-distillation variants~\citep{yang2026self,zhao2026self} for token supervision.
These works primarily focus on credit assignment on tokens within individual policy updates, while we instead allocate credit over the longtime training process, that orchestrating the optimization.

\textbf{Token-selective optimization.}
Recent analyses of reinforcement learning on LLMs show that the effective learning signals are highly non-uniform across tokens, with only a small fraction of tokens~\citep{wang2025beyond,meng2026sparse,chen2025reshaping} disproportionately dominating the gradient updates, which have been broadly verified through token-level divergence between the base and RL policies~\citep{huang2026direction}, ranking shifts statistics~\citep{huan2025does,chen2025reshaping}, and cross sampling performance~\citep{meng2026sparse}.
Steering RLVR optimization with minority high-entropy tokens~\citep{wang2025beyond} or divergence-weighted token-level advantages~\citep{meng2026sparse,yang2025not} induces superior performance over uniform full-token optimization.
And recent works further extend these prioritized token optimization through teacher-student divergence to on-policy distillation~\citep{kim2026explain,xu2026tiptokenimportanceonpolicy}.
These approaches primarily investigate where the learning signal lies, and implicitly assuming that all selected tokens can be optimized simultaneously, while we focus on a complementary but underexplored question: how optimization over tokens should be orchestrated across context.
\section{Conclusion}
\label{sec:conclusion}
In this work, we revisit the RLVR optimization from a temporal perspective, and shown that prevalent credit allocation methods principally suffered from stagnant allocation criteria throughout training, advocating that \textit{when} learning signals are scheduled can be as important as \textit{where} they are allocated over tokens.
Build upon this, we introduce the temporal scheduling and identify that the trajectory percentiles serves as a simple but effective perspective for evolving policy optimization.
While the framework is further extended to common credit allocation methods, including continuous advantage reweighting, and discrete sparse token optimization.
We formed the temporal scheduling by prioritizing targeted tokens that emphasized by respective allocation criteria, and gradually attenuating toward general optimization.
And we show that the sacrificed policy entropy is largely reserved by temporally scheduling the policy optimization.
Experiments across model scales and RLVR algorithms demonstrate consistent improvements, suggesting a promising optimization dimension.
Future works involve the orchestration of diverse credit allocation criteria in optimization.

\bibliography{main}
\bibliographystyle{unsrtnat}

\clearpage
\appendix 
\section{Implementation Details}
\label{sec:appendix_implementation}
\subsection{Training Configurations}
\label{sec:app-detail}
We detail the training configurations and temporal scheduling hyperparameters employed throughout experiments in Table~\ref{tab:training_config}.
We conduct experiments on 8 H200 GPUs, and uses AdamW optimizer with weight decay of $0.1$.
The learning rate follows a constant schedule.
Gradient clipping is applied with maximum gradient norm set to $1.0$.
During training, we use 2048 sampled trajectories for policy optimization at each update step, and perform on-policy rollout generation throughout training.

For temporal scheduling, we adopt the linear schedule as the default setting in the main experiments, while additionally exploring sigmoid and gamma scheduling variants.
We use trajectory percentile as default credit allocation setting for temporally scheduling the policy optimization, and starts from the strongest credit allocation criteria with $\tau_{low}=0$, and apply the maximal temporal range to $\tau_{high}=0$.

\begin{table}[!htb]
\caption{Training hyperparameters and temporal scheduling configurations used in our experiments, where \underline{underline} denotes the default settings with variants for presentation}
\centering
\small
\setlength{\tabcolsep}{6pt}
\renewcommand{\arraystretch}{1.08}
\begin{tabular}{l|c}
\toprule
\textbf{Hyperparameter} & \textbf{Temporal Scheduling} \\
\midrule
\multicolumn{2}{c}{\textit{Shared Optimization Settings}} \\
\midrule
Base Model & Qwen3-4B / Qwen3-8B \\
Global Batch Size & 256 \\
Group Rollout Size & 8 \\
Optimizer & AdamW \\
Learning Rate & $1\times10^{-6}$ \\
LR Scheduler & Constant \\
Weight Decay & 0.01 \\
Gradient Clipping & 1.0 \\
Temperature & 1.2 \\
Max Prompt Length & 1,024 \\
Max Response Length & 30,996 \\
Policy Clip Ratio & [0.2,\,0.28] \\
\midrule
\multicolumn{2}{c}{\textit{Temporal Scheduling Settings}} \\
\midrule
Scheduling Function $\mathcal{S}(\cdot)$& \underline{Linear} / Sigmoid / Gamma \\
Sigmoid Schedule Center ($\beta$, $k$) & (0.5,\,8) \\
Gamma Schedule Factor ($\gamma$) & 2.0 \\
Temporal Range ($\tau_{low}, \tau_{high}$) & (0,\,0.8) \\
Credit Allocation Criteria & Entropy / \underline{Trajectory Percentile} \\
\bottomrule
\end{tabular}
\label{tab:training_config}
\end{table}

\subsection{Scheduling Functions.}
\label{sec:schedule_function}
We explore the following scheduling functions $\mathcal{S}(\cdot)$ for progressively expanding optimization throughout training. 
Recall that all scheduling functions are monotonically decreasing with respect to normalized training progress $\bm{\tau} \in [0,1)$, and subject to $\mathcal{S}(\bm{\tau})\in[0,1)$.
Larger scheduling values correspond to stricter optimization selection during early training, while smaller values gradually expand optimization toward broader sampled tokens in later stages.

\begin{itemize}[leftmargin=1.2em,itemsep=0.2em]
    \item \textsc{Linear Schedule:}
    $\mathcal{S}(\tau)=1-\tau$.
    The optimization scope increases linearly along training progress.

    \item \textsc{Sigmoid Schedule:}
    $\mathcal{S}_{\mathrm{sigmoid}}(\tau)=\frac{1}{1+\exp\big(k(\tau-\beta)\big)}$.
    The optimization expansion remains conservative during early training and later training, and accelerates near the middle stage. We adopt $k=8$ to control the sharpness of the transition, while $\beta=0.5$ to specify the transition center along the training progress.

    \item \textsc{Gamma Schedule:}
    $\mathcal{S}(\tau)=(1-\tau)^\gamma$.
    Larger $\gamma$ delays optimization expansion toward later training stages, while smaller $\gamma$ yields more aggressive early inclusion, and we use $\gamma=2$ in our experiments.
\end{itemize}

\subsection{Credit Allocation Proxies}
\label{sec:allocation_proxy}
Temporal scheduling operates on token-level credit allocation proxies that determine the contribution of each sampled tokens for policy optimization.
Given sampled trajectory $\mathbf{y}=\{y_1,\dots,y_T\}$, each token $y_t$ is associated with corresponding derived proxy $\mu_t$, and we primirally focus on discrete case described in Sec~\ref{sec:method:schedule}, where the scheduled allocation factor is generally formulated as:
\[
\widehat{f}_{t(\tau)}
=
\mathbb{I}
\big[
\mu_t \ge \mathcal{S}(\tau)\cdot \epsilon
\big],
\]
where $\mathcal{S}(\tau)$ denotes the temporal scheduling function and $\epsilon$ controls the optimization boundary.

\begin{itemize}[leftmargin=1em,itemsep=0.35em]

\item \textsc{Trajectory Percentile Proxy:}
$\mu_t = \frac{t}{T}\in[0,1]$.
Larger $\mu_t$ corresponds to tokens located closer to the end of the trajectory.
The scheduled allocation factor is defined as $\widehat{f}_{t(\tau)}=\mathbb{I}[\mu_t \ge \mathcal{S}(\tau)]$ for suffix optimization, which progressively extends optimization from later toward earlier tokens throughout training, and $\widehat{f}_{t(\tau)}=\mathbb{I}[\mu_t \le \mathcal{S}(1-\tau)]$ for prefix optimization, as described in Figure~\ref{fig:ablation_proxy}.

\item \textsc{Entropy-based Proxy:}
$\mu_t=\mathcal{H}(\pi_\theta(\cdot\mid y_{<t}))$.
The scheduled allocation factor is defined as $\widehat{f}_{t(\tau)}=\mathbb{I}[\mu_t\ge \mathcal{S}(\tau)\cdot\epsilon]$, where typically $\epsilon=0.2$ without scheduling~\citep{wang2025beyond}, and we set $\epsilon=1$ for temporal scheduling. Optimization is initialized from most high-entropy tokens and progressively expanded toward broader sampled tokens during training.

\item \textsc{Random Proxy:}
$\mu_t\sim\mathrm{Bernoulli}(\bm{\tau})$, where the scheduled progress $\bm{\tau}$ controls the selected probability of sampled tokens. 
The scheduled allocation factor is defined as $\widehat{f}_{t(\tau)}=\mathbb{I}[\mu_t \ge \mathcal{S}(\tau)]$.
Unlike structured scheduling criteria, random token selection does not preserve coherent policy behaviors along the trajectory and typically results in unstable optimization.

\end{itemize}
\section{Further Analysis on Temporal Scheduling}
\begin{figure*}[!htb]
    \centering
    \includegraphics[width=0.99\linewidth]{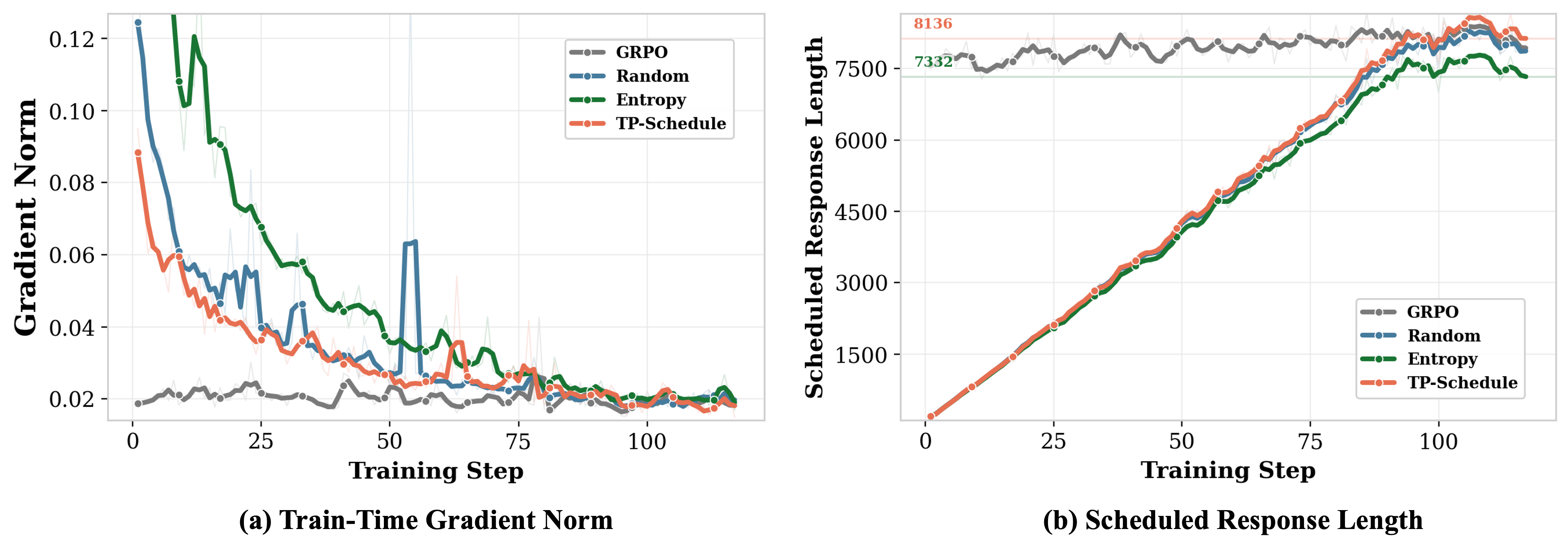}
    \vspace{-1em}
    \caption{Training dynamics under temporal scheduling. (a) Train-time gradient norm. Temporal scheduling consistently yields larger and more concentrated optimization signals compared to standard GRPO, while TP-Schedule exhibits the most stable optimization trajectory. Random scheduling introduces unstable gradient spikes, suggesting that incoherent token organization corrupts the accumulated policy gradients. (b) Scheduled response length during optimization. Temporal scheduling progressively expands the optimization from specified token subsets toward broader trajectory, where TP-Schedule achieves slightly higher response length compared to entropy-based schedule.}
    \label{fig:grad_length}
    \vspace{-1em}
\end{figure*}

We further analyze the optimization dynamics induced by temporal scheduling through gradient norm and scheduled response length statistics, as shown in Figure~\ref{fig:grad_length}.

\textbf{Gradient Norm Dynamics.}
We first examine the train-time gradient norm in Figure~\ref{fig:grad_length} (a). Compared to standard GRPO, temporal scheduling methods exhibit significantly larger gradient norms during the early training stage, indicating that optimization is concentrated on more informative token subsets rather than uniformly averaged across all sampled tokens.
Among different scheduling strategies, TP-Schedule shows the most stable optimization trajectory, where the gradient norm gradually decreases as optimization is progressively expanded toward broader trajectory positions.
While entropy-based scheduling maintains relatively strong optimization signals during early training, with higher gradient norm compared to TP-Schedule, consistent with its focus on high-uncertainty policy behaviors.
In contrast, random scheduling produces unstable optimization dynamics with significant gradient spikes during training, suggesting that randomly organizing token participation disrupts coherent policy behaviors and corrupts the accumulated policy gradients.
Meanwhile, standard GRPO maintains consistently small gradient norms due to uniformly averaging optimization over all sampled tokens, potentially diluting informative learning signals.

\textbf{Scheduled Response Length Dynamics.}
We further analyze the scheduled response length in Figure~\ref{fig:grad_length} (b), which measures the average number of trajectory tokens participating in optimization throughout training, and performed under a linear schedule.
All temporal scheduling methods progressively increase the optimization scope during training, gradually extending optimization from specified token subsets toward broader trajectory regions.
Among them, TP-Schedule shows the largest final response length, indicating that trajectory percentile provides a coherent criterion for accommodating policy behaviors in optimization with less sacrificed policy entropy.
Entropy scheduling shows comparatively slower expansion and shorter scheduled response lengths, suggesting that policy-entropy remain  sacrificed during the early training.
Interestingly, although temporal scheduling initially constrains optimization to partial trajectory regions, all scheduling variants naturally approach the response length of standard GRPO during later training stages, suggesting that temporal scheduling does not fundamentally alter the final response length distribution, but instead progressively reorganizes how policy behaviors are incorporated throughout optimization.
\section{Case Study}
\label{sec:case study}
We further present qualitative case studies to analyze how temporal scheduling affects the reasoning behaviors learned during RLVR optimization.

\definecolor{lavender}{RGB}{200, 190, 230}
\definecolor{lightblue}{RGB}{190, 210, 230} 
\begin{tcolorbox}[
    title=Question and Ground Truth,
    colback=white,
    colframe=gray,
    coltitle=white,
    fonttitle=\bfseries,
    arc=1mm,
    boxrule=0.6mm,
    left=1.5mm,   
    right=1.5mm,  
    top=1mm,     
    bottom=1mm, 
]
\textcolor{gray!70!black}{\textbf{Question:}} The parabola with equation $y=x^{2}-4$ is rotated $60^{\circ}$ counterclockwise around the origin. The unique point in the fourth quadrant where the original parabola and its image intersect has $y$-coordinate $\frac{a-\sqrt{b}}{c}$, where $a$, $b$, and $c$ are positive integers, and $a$ and $c$ are relatively prime. Find $a+b+c$.

\textcolor{gray!70!black}{\textbf{Ground Truth:}} 62
\end{tcolorbox}

\begin{tcolorbox}[
    title=Response from Baseline,
    colback=white,
    colframe=lightblue,
    coltitle=black,
    fonttitle=\bfseries,
    arc=1mm,
    boxrule=0.6mm,
    left=1mm,    
    right=1mm,   
    top=1mm,     
    bottom=1mm,  
]

To solve the problem, we are given a parabola defined by the equation:
\[
y=x^2-4
\]

and we are to rotate this parabola \textbf{60$^\circ$ counterclockwise}
around the origin. The task is to find the $y$-coordinate
of the unique point in the fourth quadrant where the original parabola
and its rotated image intersect, expressed in the form:
\[
\frac{a-\sqrt b}{c}
\]

where $a,b,c$ are positive integers, and $a$ and $c$ are relatively prime.
Then, we are to find $a+b+c$.

\vspace{0.5em}

\textbf{Step 1: Rotation of a Point}

A point $(x,y)$ rotated counterclockwise by $\theta$ becomes:
\[
X=x\cos\theta-y\sin\theta
\]
\[
Y=x\sin\theta+y\cos\theta
\]

For $\theta=60^\circ$, we have:
\[
\cos60^\circ=\frac12,
\qquad
\sin60^\circ=\frac{\sqrt3}{2}
\]
\end{tcolorbox}

\begin{tcolorbox}[
    title=Response from Baseline,
    colback=white,
    colframe=lightblue,
    coltitle=black,
    fonttitle=\bfseries,
    arc=1mm,
    boxrule=0.6mm,
    left=1mm,    
    right=1mm,   
    top=1mm,     
    bottom=1mm,  
]

So the rotation becomes:
\[
X=\frac12x-\frac{\sqrt3}{2}y
\qquad
Y=\frac{\sqrt3}{2}x+\frac12y
\]

\vspace{0.5em}
\textbf{Step 2: Expressing the Rotated Parabola}

Since the original parabola is $y=x^2-4$, any point $(x,y)$ on the original
parabola satisfies this. When this point is rotated, the resulting point
$(X,Y)$ lies on the rotated parabola. However, we are interested in the
intersection of the original and rotated parabola, so we want to find
a point $(x,y)$ such that:
\begin{itemize}
    \item $y=x^2-4$ (original parabola), and
    \item $(X,Y)=\left(
    \frac12x-\frac{\sqrt3}{2}y,
    \frac{\sqrt3}{2}x+\frac12y
    \right)$ lies on the original parabola,
    i.e., $Y=X^2-4$
\end{itemize}

So, we substitute $y=x^2-4$ into the rotated point equations and set
$Y=X^2-4$. This gives an equation in terms of $x$, which we solve to find
the x-coordinates of the intersection points.

After substitution and simplification, the resulting equation is a quartic in $x$:
\[
\color{red}{
3x^4-2\sqrt3x^3-25x^2+6\sqrt3x+40=0
}
\]
{\footnotesize
\textcolor{red!70!black}{
Annotation: The intersection condition is formulated with the wrong rotation direction.
For the counterclockwise rotated parabola, the correct quartic should be
\(
3x^4+2\sqrt3x^3-25x^2-6\sqrt3x+40=0
\), where the signs of the cubic and linear terms are reversed.}}

\vspace{0.5em}

\textbf{Step 3: Solving the Quartic Equation}

By substituting $z=x\sqrt3$, the equation becomes:
\[
z^4-2z^3-25z^2+18z+120=0
\]

This factors as:
\[
(z^2+z-10)(z^2-3z-12)=0
\]

Solving these quadratics gives:
\[
z=\frac{-1\pm\sqrt{41}}{2}
\qquad
z=\frac{3\pm\sqrt{57}}{2}
\]

Then, since $z=x\sqrt3$, we get:
\[
x=\frac{-1\pm\sqrt{41}}{2\sqrt3},
\qquad
x=\frac{3\pm\sqrt{57}}{2\sqrt3}
\]

\vspace{0.5em}

\textbf{Step 4: Identify the Point in the Fourth Quadrant}

We are looking for the unique point in the fourth quadrant,
which requires:
\begin{itemize}
    \item $x>0$ 
    \item $y<0$
\end{itemize}

Let’s test the values:
\begin{itemize}
    \item For
    \[
    x=\frac{-1+\sqrt{41}}{2\sqrt3}
    \approx
    \frac{-1+6.4}{3.464}
    \approx 1.56,
    \]
    positive, and
    \[
    y=x^2-4\approx2.43-4=-1.57,
    \]
    negative $\rightarrow$ this is a valid point in the fourth quadrant.
\end{itemize}

But when we compute the y-coordinate of this point, it turns out to be:
\[
y=\frac{-3-\sqrt{41}}{6}
\]

This is not in the form $\frac{a-\sqrt b}{c}$, as the numerator is
negative and contains a negative sign.

\end{tcolorbox}

\begin{tcolorbox}[
    title=Response from Baseline,
    colback=white,
    colframe=lightblue,
    coltitle=black,
    fonttitle=\bfseries,
    arc=1mm,
    boxrule=0.6mm,
    left=1mm,    
    right=1mm,   
    top=1mm,     
    bottom=1mm,  
]

However, if we consider the {\color{red} clockwise rotation} instead,
the correct solution emerges:
\begin{itemize}
    \item The correct point is
    \[
    \color{red}{
    x=\frac{-3+\sqrt{57}}{2\sqrt3}
    }
    \]
    {\footnotesize
\textcolor{red!70!black}{Annoation: This solution is introduced by switching to a clockwise rotation, which is inconsistent with the original problem statement specifying a counterclockwise rotation.}}
    \item Then,
    \[
    y=x^2-4=\frac{3-\sqrt{57}}{2}
    \]
\end{itemize}

This is in the form $\frac{a-\sqrt b}{c}$, where:
\begin{itemize}
    \item $a=3$
    \item $b=57$
    \item $c=2$
\end{itemize}

These are all positive integers, and $\gcd(3,2)=1$,
so the condition is satisfied.

\vspace{0.5em}

\textbf{Final Step: Compute $a+b+c$}

\[
a+b+c=3+57+2=\boxed{62}
\]

\begin{center}
\color{red!65!black}{Correct Answer with Incorrect Reasoning Process}
\end{center}
\end{tcolorbox}

\begin{tcolorbox}[
    title=Response from Temporal Scheduling,
    colback=white,
    colframe=lavender,
    coltitle=black,
    fonttitle=\bfseries,
    arc=1mm,
    boxrule=0.6mm,
    left=1mm,    
    right=1mm,   
    top=1mm,     
    bottom=1mm,  
]

To solve the problem, we are given that the parabola $y=x^2-4$ is rotated $60^\circ$ counterclockwise about the origin, and we are to find the unique
point in the fourth quadrant where the original and rotated parabolas intersect.
The y-coordinate of this point is given in the form
\[
\frac{a-\sqrt b}{c},
\]
and we are to find $a+b+c$, where $a$ and $c$ are coprime.

\vspace{0.5em}

\textbf{Step 1: Understand the Rotation}

A rotation of a point $(x,y)$ by $60^\circ$ counterclockwise around the origin is given by
the rotation matrix:
\[
\begin{bmatrix}
x' \\
y'
\end{bmatrix}
=
\begin{bmatrix}
\cos 60^\circ & -\sin 60^\circ \\
\sin 60^\circ & \cos 60^\circ
\end{bmatrix}
\begin{bmatrix}
x \\
y
\end{bmatrix}
=
\begin{bmatrix}
\frac12 & -\frac{\sqrt3}{2} \\
\frac{\sqrt3}{2} & \frac12
\end{bmatrix}
\begin{bmatrix}
x \\
y
\end{bmatrix}
\]

So, the rotated coordinates $(X,Y)$ of a point $(x,y)$ are:
\[
X=\frac12x-\frac{\sqrt3}{2}y,
\qquad
Y=\frac{\sqrt3}{2}x+\frac12y
\]

We are to find the equation of the rotated parabola, which is the image of the original
parabola under this rotation.

\vspace{0.5em}

\textbf{Step 2: Find the Equation of the Rotated Parabola}

We solve the system:
\[
X=\frac12x-\frac{\sqrt3}{2}y
\qquad
Y=\frac{\sqrt3}{2}x+\frac12y
\]

Solving for $x$ and $y$ in terms of $X$ and $Y$, we get:
\[
x=\frac{X+\sqrt3Y}{2},
\qquad
y=\frac{Y-\sqrt3X}{2}
\]
\end{tcolorbox}

\begin{tcolorbox}[
    title=Response from Temporal Scheduling,
    colback=white,
    colframe=lavender,
    coltitle=black,
    fonttitle=\bfseries,
    arc=1mm,
    boxrule=0.6mm,
    left=1mm,    
    right=1mm,   
    top=1mm,     
    bottom=1mm,  
]
Since the original parabola is $y=x^2-4$, substitute these expressions into it:
\[
\frac{Y-\sqrt3X}{2}
=
\left(
\frac{X+\sqrt3Y}{2}
\right)^2-4
\]

Multiply both sides by 4:
\[
2(Y-\sqrt3X)=(X+\sqrt3Y)^2-16
\]

Expanding the right-hand side:
\[
(X+\sqrt3Y)^2
=
X^2+2\sqrt3XY+3Y^2
\]

So the equation becomes:
\[
2Y-2\sqrt3X
=
X^2+2\sqrt3XY+3Y^2-16
\]

Bring all terms to one side:
\[
X^2+2\sqrt3XY+3Y^2+2\sqrt3X-2Y-16=0
\]

This is the equation of the rotated parabola in the original coordinate system.

\vspace{0.5em}

\textbf{Step 3: Find the Intersection with the Original Parabola}

We are to find the points of intersection between the original parabola
$y=x^2-4$ and the rotated one. Substituting $y=x^2-4$ into the rotated parabola equation:
\[
x^2+2\sqrt3x(x^2-4)+3(x^2-4)^2+2\sqrt3x-2(x^2-4)-16=0
\]

Simplify each term:
\begin{itemize}
    \item $2\sqrt3x(x^2-4)=2\sqrt3x^3-8\sqrt3x$
    \item $3(x^2-4)^2=3(x^4-8x^2+16)=3x^4-24x^2+48$
    \item $-2(x^2-4)=-2x^2+8$
\end{itemize}

Putting it all together:
\[
x^2+2\sqrt3x^3-8\sqrt3x+3x^4-24x^2+48+2\sqrt3x-2x^2+8-16=0
\]

Combine like terms:
\begin{itemize}
    \item $x^4$: $3x^4$
    \item $x^3$: $2\sqrt3x^3$
    \item $x^2$: $x^2-24x^2-2x^2=-25x^2$
    \item $x$: $-8\sqrt3x+2\sqrt3x=-6\sqrt3x$
    \item Constants: $48+8-16=40$
\end{itemize}

So the equation becomes:
\[
3x^4+2\sqrt3x^3-25x^2-6\sqrt3x+40=0
\]

\vspace{0.5em}

\textbf{Step 4: Solve the Quartic Equation}

This is a quartic equation with irrational coefficients. We can factor it using substitution
or by trial. Let’s use the substitution
\[
x=\frac{\sqrt3t}{2},
\]
which simplifies the equation.

After substitution and factoring, we find that the equation factors as:
\[
(3t^2+3t-4)(9t^2-3t-10)=0
\]
\end{tcolorbox}

\begin{tcolorbox}[
    title=Response from Temporal Scheduling,
    colback=white,
    colframe=lavender,
    coltitle=black,
    fonttitle=\bfseries,
    arc=1mm,
    boxrule=0.6mm,
    left=1mm,    
    right=1mm,   
    top=1mm,     
    bottom=1mm,  
]
Solving each quadratic:
\begin{itemize}
    \item $3t^2+3t-4=0 \Rightarrow t=\frac{-3\pm\sqrt{57}}{6}$
    \item $9t^2-3t-10=0 \Rightarrow t=\frac{1\pm\sqrt{41}}{6}$
\end{itemize}

Now, recall that
\[
x=\frac{\sqrt3t}{2},
\]

so:
\begin{itemize}
    \item $x=\frac{\sqrt3(-3+\sqrt{57})}{12}=\frac{-\sqrt3+\sqrt{19}}{2}$
    \item $x=\frac{\sqrt3(1+\sqrt{41})}{12}$
    \item etc.
\end{itemize}

Only the first solution gives a point in the fourth quadrant, where
$x>0$ and $y<0$. Let’s take:
\[
x=\frac{-\sqrt3+\sqrt{19}}{2}
\]

Then, the y-coordinate is:
\[
y=x^2-4
=
\left(
\frac{-\sqrt3+\sqrt{19}}{2}
\right)^2-4
\]

Simplify:
\[
x^2
=
\frac{19+3-2\sqrt{57}}{4}
=
\frac{22-2\sqrt{57}}{4}
=
\frac{11-\sqrt{57}}{2}
\]

\[
y
=
\frac{11-\sqrt{57}}{2}-4
=
\frac{11-\sqrt{57}-8}{2}
=
\frac{3-\sqrt{57}}{2}
\]

\vspace{0.5em}

\textbf{Step 5: Final Answer}

The y-coordinate is:
\[
\frac{3-\sqrt{57}}{2}
\]

So,
\[
a=3,
\qquad
b=57,
\qquad
c=2,
\]
and:
\[
a+b+c=3+57+2=\boxed{62}
\]
\end{tcolorbox}

Although standard GRPO is able to improve final-answer accuracy through outcome-level supervision, we observe that its reasoning process often fails to establish reliable intermediate reasoning behaviors.
In cases, the generated response arrives at the correct answer while containing inconsistent, incomplete, or logically incorrect intermediate derivations.
While temporal scheduling is more able to produces coherent reasoning trajectories, suggesting that the process-level reasoning quality is also implicitly promoted during training.
We believe this phenomenon arises from the progressive organization of policy behaviors throughout optimization.
\begin{itemize}[leftmargin=1.2em,itemsep=0.2em]
    \item Standard GRPO uniformly optimizes all sampled tokens under globally broadcast rewards, which may simultaneously mix heterogeneous behaviors, encouraging shortcuts that directly improve outcome rewards without consistently refining intermediate reasoning steps.
    \item Temporal scheduling leverages the progressive optimization, avoiding simultaneously balancing heterogeneous policy behaviors during early training, allowing intermediate reasoning structures to emerge more consistently, suggesting a promising dimension to complement existing credit allocation methods for reasoning process optimization under sparse outcome supervision.
\end{itemize}

\end{document}